\documentclass{article}

\usepackage[dblblindworkshop,final]{neurips_2026}
\workshoptitle{SLMs for Agentic Systems, Paris, France, 2026}

\usepackage[utf8]{inputenc} 
\usepackage[T1]{fontenc}    
\usepackage{hyperref}       
\usepackage{url}            
\usepackage{booktabs}       
\usepackage{amsfonts}       
\usepackage{nicefrac}       
\usepackage{microtype}      
\usepackage{xcolor}         

\usepackage{graphicx}
\usepackage{makecell}
\usepackage{multirow}
\usepackage{amsmath,amssymb,bm}
\usepackage{dsfont}                        
\newcommand{\mathbbm}[1]{\mathds{#1}}      

\usepackage{listings}
\usepackage[most]{tcolorbox}
\tcbuselibrary{skins,breakable,listings}

\definecolor{promptbg}{HTML}{F8F8FA}
\definecolor{promptframe}{HTML}{2F4F7F}
\definecolor{promptph}{HTML}{A03030}   

\lstdefinestyle{promptstyle}{
  basicstyle=\ttfamily\scriptsize,
  breaklines=true,
  breakautoindent=false,
  breakindent=0pt,
  columns=fullflexible,
  keepspaces=true,
  showstringspaces=false,
  upquote=true,
  extendedchars=true,
  xleftmargin=0pt,
  aboveskip=0pt, belowskip=0pt,
  alsoletter={_},
  morekeywords={TOOL_SIGNATURES_AND_EXAMPLES,TOOL_NAME,JSON_SCHEMA_OF_TOOL,
                UP_TO_N_SEED_EXAMPLES_AS_JSON,DIMENSIONS,NUM_DIMENSIONS,
                COMMA_SEPARATED_TOOL_NAMES,LIST_OF_DIMENSION_VALUE_LISTS},
  keywordstyle=\color{promptph}\bfseries,
  literate=
    {⊆}{{$\subseteq$}}1
    {∪}{{$\cup$}}1
    {−}{{$-$}}1
    {×}{{$\times$}}1
    {→}{{$\rightarrow$}}1
    {—}{{---}}1
    {’}{{'}}1
    {“}{{``}}1
    {”}{{''}}1
}

\newtcolorbox[auto counter, list inside=prompts]{promptboxenv}[2][]{%
  breakable, enhanced,
  colback=promptbg,
  colframe=promptframe,
  coltitle=white,
  fonttitle=\bfseries\small,
  title={Prompt~\thetcbcounter: #2},
  boxrule=0.6pt,
  arc=2pt,
  left=4pt, right=4pt, top=4pt, bottom=4pt,
  #1%
}

\newcommand{\promptbox}[3][]{%
  \begin{promptboxenv}[#1]{#2}%
    \lstinputlisting[style=promptstyle]{#3}%
  \end{promptboxenv}%
}

\title{Turnslide: Scalable Multi-Turn Data Synthesis by Walking a Finite-State Machine}

\author{%
  Aaron Fainman \\
  Imperial College London \\
\And
  Gabriela Kadlecová \\
  distil labs \\
\And
  Maciej Gryka \\
  distil labs \\
\And
  Bartosz Kruszczyński \\
  distil labs \\
\And
  Usman Zafar \\
  distil labs \\
\And
  Cédric Archambeau \\
  Agon \\
\And
  Aaron Klein \\
  ELLIS Institute Tübingen \\
\And
  David Salinas \\
  ELLIS Institute Tübingen \\
\And
  Selim Nowicki \\
  distil labs \\
\And
  Jacek Golebiowski \\
  distil labs \\
}

\begin{document}

\maketitle

\begin{abstract}
Small language models are inexpensive to serve and can run on private infrastructure, but base models are often not good enough at multi-turn tool calling, and fine-tuning them needs per-API data that rarely exists. Existing synthesis methods are too expensive for high-scale fine-tuning, as they often require mock operational environments for different domains and multiple LLM calls per generated conversation turn. We introduce a fully automated, lightweight synthesis framework that models each API as a finite-state machine, representing the system as abstract states that determine when each tool may be called, producing state-valid sequences of tools; sequences are translated into complete examples with a single LLM call. Rather than optimize diversity, we set a target distribution over the number of turns, the tool sequence and task complexity. We measure data quality by fine-tuning SLMs on generated trajectories, showing that our FSM-based generation significantly improves downstream accuracy over an unmutated baseline and, against existing works, reaches 70.7\% full accuracy over 63.4\% and 53.7\% with 3.6--6.6$\times$ fewer tokens.
\end{abstract}

\section{Introduction}

Small language models (SLMs) are inexpensive to serve, respond with low latency, and are small enough to run on a user's own infrastructure, which matters wherever data cannot leave the environment. Tool calling extends such models beyond text generation, letting them act on external systems and databases on a user's behalf~\citep{toolformer, toolbench}. In the multi-turn setting, the assistant fulfills a task over a sequence of interactions with a human user, issuing tool calls and incorporating their responses until the task is complete.

Unfortunately, base SLMs are not competitive at this task. In the multi-turn category of the Berkeley Function Calling Leaderboard (BFCL~v4)~\citep{Patil25} even frontier models are lacking, with Gemini-3-Pro~\citep{gemini3-pro} scoring only 61\% and GPT-5.2~\citep{gpt5} 44\% on accuracy. Untuned SLMs collapse on it: Llama-3.2-3B-Instruct~\citep{llama3.2-3b}, one of the students we fine-tune, scores 4.0\% on the same category, Qwen3-1.7B 11.0\%, and Llama-3.2-1B-Instruct 0.0\%~\citep{bfclresults}. Fine-tuning closes the gap - at the same 3B scale the specialized xLAM-2-3b-fc-r reaches 58.4\%, and xLAM-2-8b-fc-r reaches 70.0\%~\citep{apigenmt}, second overall on the category and ahead of every frontier model evaluated, which shows that size alone does not solve the task but specialized training does~\citep{belcak2025}.

However, fine-tuned SLMs are not trivial to build: every new API requires training data that interacts with it, and such data rarely exists. Public benchmarks cover a small number of domains, human annotation of multi-turn trajectories is slow and costly, and the data must be produced again whenever the tool schema changes~\citep{apigenmt}. Synthetic generation is the practical route~\citep{toolbench, liu24, toolace, magnet}, and it has to satisfy two conditions: it must be inexpensive enough to run for a single task, and the data it produces must be good enough to train a model worth deploying.

Existing synthesis methods for multi-turn tool calling fall short on both conditions. Generation is typically preceded by a structural blueprint, fixing attributes such as user intent, persona, and the sequence of tools to call~\citep{apigenmt, button, toolflow}. The conversation is then generated turn by turn, with LLMs roleplaying the user, the assistant, and the tool responses, in a process termed dialogue-guided synthesis~\citep{button, toolflow, apigenmt, tooldial, orchdag, genesisfunc}. Each role requires a separate LLM invocation, which means every single trajectory requires tens~\citep{toolflow} to hundreds of calls~\citep{apigenmt} which amounts to hundreds of thousands of tokens for a single rollout. This makes such synthesis out of reach for applications where models need to be built often. Tool responses may alternatively be produced by executing calls against live endpoints or mock operational sandboxes~\citep{toolbench, apigenmt, stabletoolbench}. This grounds the trajectory in real system behavior, but the sandbox has to be built by hand for each new API, which limits scalability.

Dataset diversity presents a second difficulty. The blueprint generally specifies the sequence of tools appearing in the trajectory, and these sequences are often sampled from a graph over tools~\citep{toolflow, tooldial, apigenmt, magnet}. Such graphs are stateless: they indicate which tools may follow one another but ignore what the conversation has already done. Sampling a path may therefore produce a trajectory that violates a system precondition, and these trajectories are discarded after generation, spending budget on samples that are never used.

In this work, we address both those issues and introduce a fully automated framework for synthesizing multi-turn tool-calling data. The framework takes a task description, tool schemas, and a small seed corpus, and returns a synthetic dataset without human authoring at any stage. We model each API as a simplified finite-state machine (FSM) that an LLM constructs. Sampling a tool sequence then becomes a walk along the FSM, constrained to valid state transitions, so that trajectories are state-consistent by construction rather than by post-hoc filtering. Because the blueprint already carries the system state, a single LLM call is enough to generate the whole conversation.

Our method helps us to increase diversity, as we can specify a target distribution over the number of turns, the tool sequence, and task complexity, and derive samplers that realize it. We also present phantom tool calls -- additional instructions that force retrieval of information or arithmetic computation before a parameter can be supplied.

To summarize our contributions:
\begin{itemize}
    \item We present a synthetic data generation method for multi-turn tool calling that models API dynamics as an automatically constructed finite-state machine, generates each trajectory in a single LLM call, and controls diversity by specifying the distribution over selected blueprint attributes.
    \item We evaluate the method by fine-tuning a student SLM on synthetically created traces and reporting its benchmark accuracy before and after training for each API domain derived from BFCL~v4~\citep{Patil25}. Training on data created with FSM-based sampling has the highest positive impact of all the methods we tested.
    \item We benchmark the method against ToolFlow~\citep{toolflow} and APIGen-MT~\citep{apigenmt} on data quality, and show that our data trains better students, reaching 70.7\% full accuracy against 63.4\% and 53.7\%, with higher judged data quality (3.34 against 3.03 and 3.04).
    \item We benchmark the same two methods on generation cost, and show that ours is significantly more efficient to run, at 1.0 LLM call per trajectory against 34.3 and 107.2, producing 3.6--6.6$\times$ more trajectories under a fixed 20M-token budget.
\end{itemize}

\section{Method}\label{sec:method}

\begin{figure}[ht]
\vskip 0.2in
\begin{center}
\centerline{\includegraphics[width=\columnwidth]{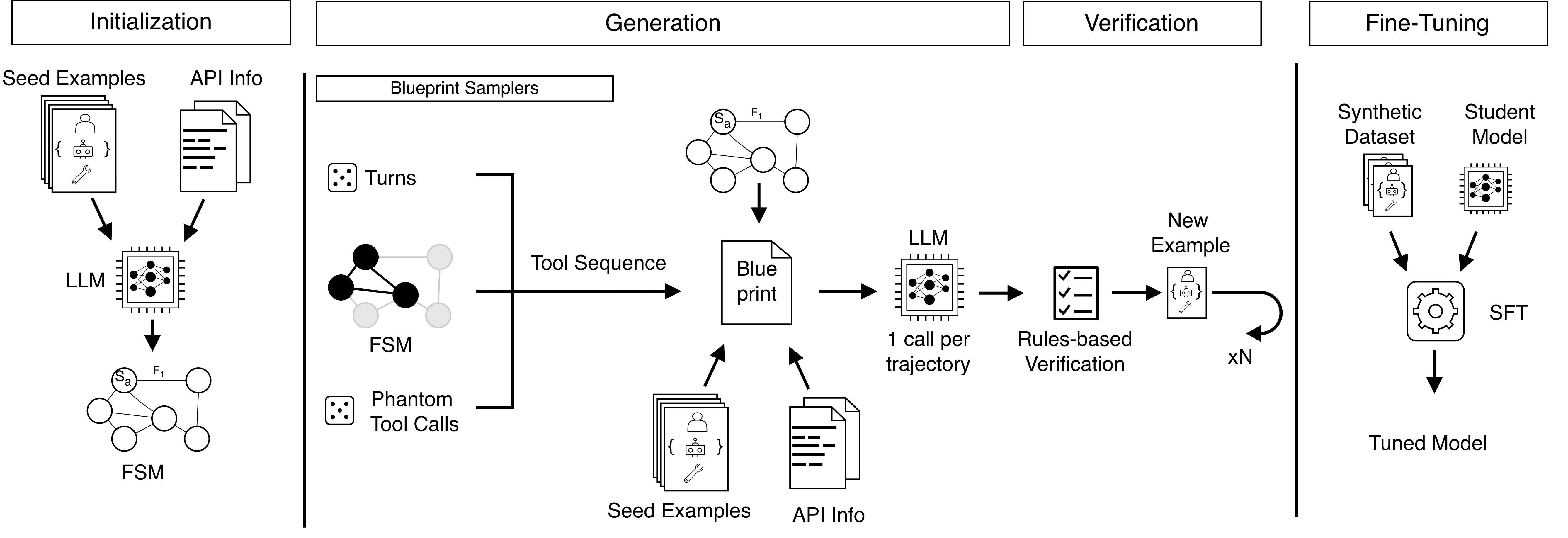}}
\caption{\textcolor{black}{ Overview of our end-to-end pipeline showing offline initialization of the FSM, assembly of the blueprint by sampling specific attributes, generation of conversation trajectories in a single pass, and verification of examples. Downstream performance is measured after fine-tuning an SLM. }}
\label{fig:pipeline}
\end{center}
\vskip -0.2in
\end{figure}

In multi-turn tool calling, the validity of a tool call often depends on an underlying system state that is not explicitly accessible. For example, updating an account balance may first require authenticating the user, and starting a vehicle may require a brake to be engaged first. We model that state, and how API calls change it, with a finite-state machine (FSM) that LLMs construct offline. Sampling from the FSM gives tool sequences that are valid by design, and lets us set a target distribution over them rather than optimize for diversity. To create valid examples, that sequence, along with seed examples, tool schemas, and the task description, forms an example blueprint from which a single LLM call synthesizes the whole conversation. Fig.~\ref{fig:pipeline} shows the pipeline.

\subsection{Finite-State Machine}

We model each API domain using a simplified finite-state machine (FSM) comprising
\begin{itemize}
    \item states ($S$): representations of the system that determine which tool invocations are valid;
    \item tool functions ($F$): the set of tools that may be invoked;
    \item guards ($G$): conditions of the system state that determine whether a tool may be called; and
    \item transition relations ($T$): rules governing how the system state changes.
\end{itemize}

\paragraph{Construction} The FSM is constructed automatically with two LLM calls. The first identifies the underlying system states from the task description, tool schemas, and seed examples. The second determines the guard and transition relations for each tool. The recovered FSM is not verified against a downstream specification of the API (see Sec.~\ref{sec:limitations}), relying instead on downstream performance as a metric for fidelity.

\paragraph{Example} As an example, consider a simplified FSM for the VehicleControlAPI in BFCL~v4~\citep{Patil25}, with three binary state variables: doors are locked, brake pedal is pressed, engine is running. Tools like \texttt{lockDoors} and \texttt{pressBrakePedal} do not have a state guard but toggle the doors being locked and the brake pressed, respectively. The \texttt{startEngine} tool is guarded on \textsc{doors\_locked} $\wedge$ \textsc{pedal\_pressed} and transitions \textsc{engine\_off} $\rightarrow$ \textsc{engine\_on} or vice-versa. A walk along the FSM therefore cannot start the engine before locking the doors and pressing the brake, whereas a stateless tool graph permits state-invalid sequences.

\paragraph{States rather than tools} Prior work samples tool sequences from graphs over tools~\citep{toolflow, tooldial, toolace, magnet,apigenmt}. These graphs are Markovian in the relationship between tools: whether a tool may be called depends only on the preceding tool. However, tool invocation generally depends on the entire conversation history and on the initial environment state. In our construction, the state transitions are Markovian, but the tool sequence is not since the tool distribution is conditioned on the current state. This allows for additional flexibility, potentially allowing tool interactions across multiple turns.

\subsection{Blueprint Sampling}\label{sec:blueprint-sampling}

A blueprint is the complete specification of one trajectory, used to synthesize a complete conversation. It is based around the sample from the FSM which fixes what the trajectory must contain before any text is written: how long it runs, which tools are called and in what order, and where extra reasoning is required. Alongside FSM states, it carries extra material to ground the synthesis: the tool schemas, the task description, and the seed examples. Constructing a blueprint is therefore sampling from the FSM. The sampling is guided by three attributes and good coverage of their respective distributions leads to high data diversity:

\textbf{Number of turns}: uniform over a fixed range (in our runs 2-10), approximately the maximum conversation length in comparable datasets~\citep{Patil25}.

\textbf{Tool sequence}: given the number of turns, the sequence is a walk along the FSM. In each state a tool is drawn from those its guard permits, and the tool's transition relation gives the next state. Drawing each permitted tool with equal probability does not make the tools uniform overall, because the guards and the transition structure visit some states far more often than others. We therefore solve for the per-state sampling probabilities that bring the marginal distribution of tools over the whole conversation as close to uniform as possible, a convex subproblem we solve with CVXPY (see App.~\ref{app:tool-sampling}).

\textbf{Task complexity}: real trajectories contain multi-step reasoning and intermediate processing rather than a bare sequence of calls. We raise complexity by inserting phantom tool calls at random positions in the sequence. These are not tools in the API. They signal to the generator that information must be retrieved or transformed before the next call can be made. We use \texttt{information\_retrieval} and \texttt{arithmetic\_computation}, placed before tools with numerical or string-valued parameters, so that \texttt{arithmetic\_computation} before a \texttt{set\_balance} operation asks for a unit conversion or a percentage change.

\subsection{Single-Pass Generation and Verification}
Our method generates each example using a single LLM invocation, based on the blueprint built in the preceding steps. This is substantially more lightweight than dialogue-guided synthesis, in which separate LLM invocations are required across conversation turns and roles. We present our data generation prompt in App.~\ref{app:prompts}. After generation, we perform a lightweight verification step. All referenced tool calls and their parameters are checked against the function signatures, while a basic ROUGE-based diversity filter discards duplicates. The rejection rate of examples is relatively low~($\approx 2.3\%$).

\section{Experiment Results}

\subsection{Experimental Setup}\label{sec:exp-setup}

To evaluate the effectiveness of our data synthesis framework, we fine-tune student models on our synthetic datasets and evaluate their downstream tool-calling performance using the Berkeley Function Calling Leaderboard (BFCL v4)~\citep{Patil25}. We adapt the benchmark into a domain-partitioned dataset by grouping examples across distinct APIs and splitting each domain into seed and test sets (50:50). The training examples form the seed corpus for our generation pipeline. Our objective is to evaluate our data synthesis method rather than to train a general tool-calling model. We therefore train, evaluate, and report performance separately for each API domain. This creates a challenging low-shot synthesis setting in which the seed corpus for a given API often contains fewer than 10 examples.

In our experiment, we expanded parallel tool calls into multiple turns with a single tool call, to help with consistent evaluation across all model families\footnote{vLLM does not support parallel tool calling for the Llama family: \url{https://docs.vllm.ai/en/v0.28.0/features/tool_calling}}. We believe the impact on post-training performance was low, as it represents a realistic scenario -- a rolled out conversation setting, where the user passes tool outputs after each call.

Our generation pipeline uses GLM-5.1~\citep{glm5} as the teacher model to synthesize 5000 synthetic examples. Downstream evaluation is performed by fine-tuning Qwen3.5-2B~\citep{qwen3.5-2b} and Llama-3.2-3B-Instruct~\citep{llama3.2-3b}, averaging across students. We perform two-fold cross-validation over the train/test partitions. After a 50:50 split on these small datasets, fewer than 10 examples remain in the test set, which is too small for stable evaluation. To strengthen the evaluation set, we augment it with additional trajectories generated using the Fable~5 model~\citep{anthropic2026fable}, generated independently of the pipeline being tested, verified by executing against the BFCL implementation, and verified by the authors. These supplementary examples are strictly held out from training and remain exclusively in the test set across both validation folds. Training details are provided in~App.~\ref{sec:training_details}.

To isolate the impact of our state-aware modeling and controlled diversity mechanisms, we compare four generation conditions:
\begin{itemize}
    \item Unmutated Baseline: Synthetic trajectories are generated directly from seed examples and task description, without any sampled tool sequences or turns specifiers.
    \item Turns: The number of turns is randomly sampled and specified in the blueprint.
    \item Markovian Tool Graph:  The blueprint specifies the number of turns, and a tool sequence from sampling a Markovian tool graph, where edge transitions are derived from co-occurring tool pairs in the seed data. Phantom tool calls are inserted into the trajectory.
    \item FSM (Ours): The blueprint specifies the number of turns, and a tool sequence via sampling our FSM. Phantom tool calls are inserted into the trajectory.
\end{itemize}
The Markov chain experiment is used to demonstrate the effectiveness of the state modeling in the FSM against a surface-level transition graph, while the turns mutator demonstrates whether downstream performance changes result from prompt diversity. Both the Markovian setup and FSM include phantom tool calls, as initial small-scale experiments showed poor performance without them.

We report the accuracy for each domain by comparing against the ground truth tool calls in the multi-turn BFCL API implementation. We measure the response per-turn, and the fully correct response (including checking the environment state) over the entire conversation trajectory. We also show a regression score to estimate the average effect of each generation condition across domains by fitting a multiple linear regression model with the generation condition, API domain, and student model as categorical control variables. The regression coefficient, $\beta$, estimates the change in downstream performance associated with the generation condition. Alongside it we have the standard error (SE) and statistical significance score ($p\leq 0.05$).

\begin{table}[t]
\centering
\small
\setlength{\tabcolsep}{3.5pt}
\begin{tabular}{l cc cc cc cc cc cc}
\toprule
\multirow{2}{*}{Method}
 & \multicolumn{2}{c}{Summary}
 & \multicolumn{2}{c}{Trading}
 & \multicolumn{2}{c}{Travel}
 & \multicolumn{2}{c}{VehicleControl}
 & \multicolumn{2}{c}{FileSystem$^1$}
 & \multicolumn{2}{c}{Tick.\&Trav.$^2$ } \\
\cmidrule(lr){2-3} \cmidrule(lr){4-5} \cmidrule(lr){6-7}
\cmidrule(lr){8-9} \cmidrule(lr){10-11} \cmidrule(lr){12-13}
 & $\beta(\pm)SE$ & $p$
 & \makecell{Per\\Turn} & Full
 & \makecell{Per\\Turn} & Full
 & \makecell{Per\\Turn} & Full
 & \makecell{Per\\Turn} & Full
 & \makecell{Per\\Turn} & Full \\
\midrule
Base Student  & ---  & ---  & 42.7 & 26.3 & 48.0 & 21.5 & 38.5 & 5.6  & 49.4 & 18.9 & 34.3 & 12.5 \\
Baseline      & ---  & ---  & 76.9 & 43.9 & 79.6 & 47.8 & 69.4 & 18.8 & 80.7 & 67.3 & 78.1 & 38.5 \\
Turns Mutator & 2.89$\pm$2.87 & 0.321 & 76.0 & 44.9 & 76.5 & 42.3 & 64.3 & 19.1 & 88.9 & 74.2 & 76.2 & 40.4 \\
Markovian     & 7.88$\pm$3.19 & 0.03 & 76.8 & 41.5 & 83.1 & 60.6 & 70.3 & 19.7 & \textbf{94.9} & \textbf{84.5} & 75.9 & 36.6 \\
FSM           & 10.1$\pm$2.62 & $<$0.01 & \textbf{81.5} & \textbf{54.2} & \textbf{84.2} & \textbf{61.5} & \textbf{72.6} & \textbf{22.5} & 94.5 & 80.2 & \textbf{78.2} & \textbf{47.2} \\
\bottomrule
\end{tabular}
    \caption{ \centering Fine-tuning performance across different generation conditions showing the per-turn accuracy (\%), the full conversation accuracy (\%), regression coefficient $\beta$ and the standard error (\%) of the generation condition and the statistical significance $p$ of the result.}
\flushleft  \hspace{10pt} \footnotesize $^1$ GorillaFileSystem API \hspace{5pt} $^2$ TicketAPI and TravelAPI
\label{tab:main-results}
\end{table}

\subsection{Main Results}
Results are shown in Table~\ref{tab:main-results} and further results in App.~\ref{app:further-results}.

\paragraph{State-Aware Sampling vs.\ Markovian Baselines}
Results indicate that our state-aware method yields a statistically significant improvement in downstream performance ($\beta=10.1\%$, $p<0.05$).  The Markovian tool model also shows improvement but to a lesser degree ($\beta=7.88\%$, $p<0.05$). The improvement from the FSM generation is particularly evident in domains with stricter tool preconditions, such as the VehicleControlAPI, TravelAPI, and TradingBot. In these domains, state requirements are more pronounced, such as ensuring locked doors and the brake pedal pressed before starting the engine, or authenticating a user prior to booking a flight. Conversely, in state-agnostic domains, like GorillaFileSystem, where operations carry trivial state dependencies, explicit state modeling achieves comparable performance to the Markovian result.

\begin{table*}[t]
\centering
\footnotesize
\setlength{\tabcolsep}{6pt}
\begin{tabular}{l c c c c c c c c}
\toprule
Method & \makecell{Calls/\\Trajectory} & \makecell{Tokens/\\Trajectory} & \makecell{Tokens/\\Turn} & \makecell{Examples/\\20M Tokens} & \makecell{Fine-Tune\\Full\\ Acc.(\%)}  & \makecell{Fine-Tune\\Per Turn\\Acc. (\%)} & \makecell{LLM\\Eval (/5)} \\
\midrule
ToolFlow & 34.3  & 99,846 & 22,327 & 200 & 63.4 & 85.7 & 3.03 \\
APIGen-MT  & 107.2 & 184,304 & 35,943 & 109 & 53.7 & 80.1 & 3.04 \\
Ours  & \textbf{1.0}   & \textbf{27,724}  & \textbf{5,545}  & \textbf{721} & \textbf{70.7} & \textbf{87.2} & \textbf{3.34} \\
\bottomrule
\end{tabular}
\caption{Cost and Performance Results compared with other Methods}
\label{tab:exps-comparison}
\end{table*}

\subsection{Comparison with Related Work}
We compare our work against two existing synthetic data generation works, ToolFlow~\citep{toolflow} and APIGen-MT~\citep{apigenmt}. We assessed both the generation cost (showing token usage and LLM calls) and fine-tuning Qwen3.5-2B on the generated datasets. Running a full evaluation is costly owing to the large computation costs of data generation for both methods, making a full evaluation, as in Table~\ref{tab:main-results} infeasible. Instead, we have run a single domain comparison on the BFCL v4 TradingBot and set a fixed generation budget of 20M tokens. Table~\ref{tab:exps-comparison} shows the computational costs and fine-tuned performance. Differences between the results here and in Table~\ref{tab:main-results} arise from the different experimental setup.

We also present the LLM-based data evaluation score from~\citep{toolflow} which scores generated trajectories on naturalness, coherence, helpfulness, and accuracy. GLM-5.1, GPT-OSS-120B, and DeepSeek v4 Pro are used to evaluate 50 of the generated examples per method. We note that the GPT-OSS-120B evaluation of our ToolFlow re-implementation yields a score of 4.05, similar to the average score of 4.3 for GPT~4 in the original work, although the GLM-5.1 and DeepSeek scores are lower.

The results in Table~\ref{tab:exps-comparison} demonstrate the improvement in scalability with our approach. As noted, our method uses single-pass generation, requiring one LLM call per conversation trajectory, in contrast to the two dialogue-guided works, which use tens or even hundreds of calls per trajectory. Under a fixed token budget, our method generates approximately 3.6--6.6$\times$ the number of examples (or 4.0--6.5$\times$ the number of turns).

\section{Conclusion}
We present a synthetic data generation method for multi-turn tool calling that models each API as a finite-state machine, used to sample tool sequences as state-valid walks and where each trajectory is generated in a single LLM call. Across five BFCL v4 domains, our state-aware approach improves downstream accuracy over an unmutated baseline. Against ToolFlow and APIGen-MT under a fixed 20M token budget, our method produces significantly more trajectories.

\subsection{Limitations}
\label{sec:limitations}
The computational costs of running data synthesis and fine-tuning constrained our comparisons with other works, as we were able to  evaluate them only under a fixed token budget on a single domain. Part of our test set was partially LLM-generated, as otherwise we would compare the methods on a very small test set. Nevertheless, the generated data was independent of our synthesis pipeline, and was both author-verified and execution-verified. Moreover, publicly released code for baseline comparisons could not be found at the time of publication. Our work has re-implemented the experimental comparisons to the best of our ability, noting that the LLM-based data evaluation of the ToolFlow method is similar to that reported by its authors for a GPT model.

Finally, our pipeline does not verify the FSM prior to synthesis (simulating a fully autonomous regime), so errors in the FSM could propagate to the generated data. Future works could address this by using ensembles of frontier models to generate and judge the FSM.

\clearpage
\bibliographystyle{plainnat}
\bibliography{References}

\newpage
\appendix
\onecolumn

\section{Related Work}

\paragraph{Attribute- and Instance-Driven Generation} A major challenge in synthetic data generation is ensuring data diversity and domain coverage. Broadly speaking, two paradigms of data synthesis exist: instance-driven and attribute-driven~\citep{Xu24} synthesis. Instance-driven methods guide the synthesis using the original labeled data as a seed corpus~\citep{Iskander24,selfinstruct}, but their overall coverage remains constrained by the diversity of the initial seed data. Conversely, attribute-driven methods condition synthesis on specific attributes of the data such as persona~\citep{Ge24}, vocabulary choice~\citep{Eldan23}, topic~\citep{Gunasekar23}, or heuristic diversity metrics like type-token ratio, Vendi scores, or neural activation patterns~\citep{chen24, li26, nguyen25}. This offers fine-grained control but requires specifying relevant attributes, which may be difficult to scale across complex domains. Separately, some works have also combined generation with iterative instruction refinement~\citep{selfinstruct, wizardlm, magpie}. Our work uses a hybrid method, where selected attributes are specified, alongside seed examples to steer the generation.

\paragraph{Blueprint Generation and Trajectory Modeling}
Data synthesis for multi-turn tool calling conditions generation on a structural blueprint of the conversation. Blueprints differ chiefly in how they represent the problem to be solved. In some cases, the trajectory model is explicit: BUTTON composes atomic sub-tasks to form compositional multi-tool problems~\citep{button}, similar to APIGen-MT’s reverse task recombination~\citep{apigenmt}. In others, the problem representation is implicit, with many works generating tool chain graphs or API context trees~\citep{toolace}. Edges between tools are typically derived from embedding similarity between tool input and output schemas~\citep{toolflow, tooldial} or from data flow dependencies \citep{apigenmt, magnet}. Sampling paths through this graph yields the tool execution chains that structure the conversation. The conversation is thus modeled as a static graph of tools. Two variants depart from this structure. OrchDAG~\citep{orchdag} inverts the dependency, specifying the graph first and synthesizing tool schemas that satisfy its edges, while at the other extreme, GenesisFunc~\citep{genesisfunc} omits an explicit model altogether. Tools are grouped as plausibly co-occurring with the exact sequence left to emerge at generation time from agent interaction. 

In each case, the conversation model is a static structure over tools alone, with no representation of the system state. Whether a tool is invoked, however, often depends on the entire history of the conversation rather than on the preceding calls. For example, an online order can only be canceled if one has been placed earlier in the conversation.

\paragraph{Dialogue-Guided Synthesis} 
Given a blueprint, most frameworks generate the conversation turn-by-turn~\citep{button, toolflow, apigenmt, tooldial, orchdag, genesisfunc}, with one or more LLMs roleplaying the user, assistant and tool responses. Each role requires distinct prompts to maintain role consistency and track stateful information. Frameworks like APIGen-MT~\citep{apigenmt} also generate multiple candidate user messages per turn, using ensembles of frontier models to judge the content best aligned with the task and conversation~\citep{apigenmt}. This is computationally intensive. Each role requires separate LLM invocations, and combined with LLM validation and candidate user response selection, means frameworks like APIGen-MT can require dozens or even hundreds of LLM calls per conversation, presenting a major bottleneck to scalability. Approaches differ in how tool responses are generated. LLM roleplay may be used in the same way that user and assistant messages are generated ~\citep{toolflow, button, toolace, orchdag}. Alternatively, they may be produced through execution using live endpoints~\citep{toolbench} or mock operational sandboxes~\citep{apigenmt, stabletoolbench}. Execution (live or mock) guarantees state fidelity across turns but introduces substantial infrastructure overhead to build a bespoke environment per domain. Existing approaches therefore sacrifice scalability for state fidelity and dialogue quality, or structural validity for lightweight generation. Our FSM abstraction balances these two qualities, constraining tool sequences to follow state-valid transitions and conditioning the generation to be state-aware.

\section{Generation Prompts}
\label{app:prompts}
 
We construct the FSM in two stages. Stage~1
(Prompt~\ref{prompt:dimensions}) infers the state dimensions from the tool
signatures and seed interactions. Stage~2 (Prompt~\ref{prompt:transitions})
is called for batches of ten tools and induces the guard and transition relation over those dimensions. Placeholders in braces are substituted at
render time.
 
\promptbox[label=prompt:dimensions]{Dimension induction (Stage 1)}{prompt_stage1.txt}
 
\promptbox[label=prompt:transitions]{Transition induction (Stage 2)}{prompt_stage2.txt}

After the FSM is created, we sample tool-call sequences that are valid walks through it and prompt a teacher model to realize each walk as a multi-turn conversation. The generation call is split into a fixed system message and a per-walk user message (Prompt~\ref{prompt:generation}), the walk itself, the tool schemas and the dimension descriptions are incorporated into the latter.
 
\promptbox[label=prompt:generation]{Conversation generation -- System prompt}{gen-prompt-system.txt}
  
\promptbox[label=prompt:generation2]{Conversation generation -- User prompt}{gen-prompt-user.txt}
\section{Tool Sampling}
\label{app:tool-sampling}

We denote the following:
\begin{itemize}
    \item $\mathcal{F}=\{F_k\}_{k=1}^{|F|}$ the set of tools available in the conversation.
    \item $\mathcal{S}=\{S_\ell\}_{\ell=1}^{|S|}$ the set of possible system states.
    \item $p_{k,\ell}=P(F_k^{(i)}\mid S_\ell^{(i)})$ the probability of calling tool $k$ at turn $i$ given that the system is in state $\ell$ at turn $i$. We take this to be stationary, i.e. $p^{(i)}_{k,\ell}=p_{k,\ell}$ for all $i$.
    \item $\beta_\ell^{(i)}$ the probability of the system being in state $\ell$ at turn $i$, collected into the vector $b^{(i)}\in\mathbb{R}^{|S|}$.
    \item $G\in\{0,1\}^{|F|\times|S|}$ the matrix of guard relations, where $G_{k,\ell}=1$ if tool $k$ is available in state $\ell$.
    \item $T\in\{0,1\}^{|F|\cdot|S|\times|S|}$ the matrix of transition relations, with block structure such that $T_{\ell|F|+k,\,m}=1$ if calling tool $k$ in state $\ell$ transitions the system to state $m$.
    \item $\widetilde{P}\in\mathbb{R}^{|F|\times|S|}$ the guard-masked tool conditional probabilities:
    \begin{equation}
    \widetilde{p}_{k,\ell}=
    \begin{cases}
        p_{k,\ell} & \text{if } G_{k,\ell}=1,\\
        0 & \text{otherwise.}
    \end{cases}
    \end{equation}
    \item $L$ the number of turns in a conversation, with $\gamma=\mathbb{E}[L]$.
    \item $\mathbbm{1}_a$ the all-ones vector of length $a$.
\end{itemize}

The system may have one or more absorbing states, from which it cannot transition out once entered. Our aim is that the expected frequency of each tool across the conversation is as close to uniform as possible, i.e. $P(F_k)=\tfrac{1}{|F|}$. Let the vector of marginal tool probabilities at turn $i$ be $z^{(i)}\in\mathbb{R}^{|F|}$, and the average marginal tool probabilities over the $L$ turns in the conversation be $\bar{z}^{(L)}\triangleq\frac{1}{L}\sum_{i=0}^{L-1}z^{(i)}$. The optimization problem to be solved is then:
\begin{align}
    \min_{\widetilde{P},\, b^{(0)} } \left\| \bar{z}^{(L)}-\frac{1}{|F|}\mathbbm{1}_{|F|} \right\|_2^2
\end{align}

In order to solve this, we will consider how the sampling probabilities of tools affect the distribution of states and how the distribution of states affects the sampling probabilities of the different tools. First, we will impose constraints for solving our system so that they are valid probability distributions:
\begin{itemize}
    \item the conditional tool probabilities sum to 1: $\mathbbm{1}^\mathsf{T}_{|F|}\widetilde{P}=\mathbbm{1}^\mathsf{T}_{|S|}$ .
    \item the initial distribution of states sums to 1: $\mathbbm{1}^\mathsf{T}_{|S|}b^{(0)} =1$.
    \item the conditional tool probabilities respect the guard: $\widetilde{P}_{k,\ell}=0$ if $G_{k,\ell}=0$.
    \item all probabilities are non-negative: $\widetilde{P}\geq0$ and $b^{(0)}\geq 0$.
\end{itemize}

Now we turn to considering the effect of the tool sampling probabilities on the state and vice versa. Consider first the probability of tool $k$ being called at turn $i$:
\begin{equation}
    P(F^{(i)}_k)=\sum_{\ell\in\mathcal{S}} P\!\left(F^{(i)}_k\mid S_\ell^{(i)}\right)P\!\left(S_\ell^{(i)}\right)
    =\sum_{\ell\in\mathcal{S}} \widetilde{p}_{k,\ell}\,\beta_\ell^{(i)}
\end{equation}
which follows from the law of total expectation. Meanwhile the state marginal for state $m$ at turn $i$ follows by summing over all tools that transition the system from any other state to state $m$:
\begin{equation}
    \beta_m^{(i)}=\sum_{j\in\mathcal{F}}\sum_{\ell\in\mathcal{S}} T_{\ell|F|+j,\,m}\,\widetilde{p}_{j,\ell}\,\beta_\ell^{(i-1)}
    \label{eq:state-recursion}
\end{equation}

We define the state transition matrix $M\in\mathbb{R}^{|S|\times|S|}$ where entry $m,\ell$ indicates the probability of transitioning from state $\ell$ to state $m$ at the next turn:
\begin{equation}
    M_{m,\ell}=P\!\left(S_m^{(i)}\mid S_\ell^{(i-1)}\right)=\sum_{j\in\mathcal{F}} T_{\ell|F|+j,\,m}\,\widetilde{p}_{j,\ell}
\end{equation}
which we note can be simplified into a matrix product:
\begin{equation}
    M=T^{\top}\operatorname{diag}\!\left(\operatorname{vec}(\widetilde{P})\right)\left(I_{|S|}\otimes\mathbbm{1}_{|F|}\right)
    \label{eq:M-vectorized}
\end{equation}
where $\otimes$ indicates the Kronecker product, $\operatorname{vec}(\cdot)$ vectorizes matrices in column-major order, $\operatorname{diag}(\cdot)$ is the diagonal operator, and $\mathbbm{1}_{|F|}$ is the all-ones vector of length $|F|$.

The summation for the state marginals in~\eqref{eq:state-recursion} becomes $\beta_m^{(i)}=M_{m,:}\,b^{(i-1)}$, or in vector form:
\begin{align}
    b^{(i)}&=M b^{(i-1)}=M^2b^{(i-2)}\nonumber \\
    &=M^{i}b^{(0)} \label{eq:b-unrolled-M}
\end{align}

The vector of tool marginals at turn $i$ is then:
\begin{align}
    z^{(i)}&=\widetilde{P}b^{(i)} \nonumber \\
    &\stackrel{\eqref{eq:b-unrolled-M}}{=}\widetilde{P}M^{i}b^{(0)}
\end{align}

For a single walk of length $L$, the average tool distribution over the walk is
\begin{equation}
    \bar{z}^{(L)}=\frac{1}{L}\sum_{i=0}^{L-1}\widetilde{P}M^{i}b^{(0)} .
\end{equation}
Averaging over the distribution of conversation lengths, and weighting each walk by its length, gives
\begin{equation}
    \bar{z}=\frac{1}{\gamma}\sum_{i=0}^{L_{\max}} w_i\,\widetilde{P}M^{i}b^{(0)},
    \qquad w_i=\Pr[L>i],
    \label{eq:zbar}
\end{equation}
where $\gamma=\mathbb{E}[L]$. For example, if the turn count is uniform over $\{2,4,6,8,10\}$ then ${w=[1,\,1,\,0.8,\,0.8,\,0.6,\,0.6,\,\dots]}$ and $\gamma=6$.

This leaves us with the optimization problem:
\begin{subequations}
\label{eq:optimization-problem}
\begin{alignat}{2}
    \min_{\widetilde{P},\,b^{(0)}}\quad & \left\|\,\bar{z}-\tfrac{1}{|F|}\mathbbm{1}_{|F|}\right\|_2^2 &&\\
    \text{s.t.}\quad
      & \bar{z}=\frac{1}{\gamma}\sum_{i=0}^{L_{\max}} w_i\,\widetilde{P}M^{i}b^{(0)} \\
    &M=T^{\top}\operatorname{diag}\!\left(\operatorname{vec}(\widetilde{P})\right)\left(I_{|S|}\otimes\mathbbm{1}_{|F|}\right), &&\\
      & \mathbbm{1}_{|F|}^{\top}\widetilde{P}=\mathbbm{1}_{|S|}^{\top}, && \label{eq:col-stoch}\\
      & \mathbbm{1}_{|S|}^{\top}b^{(0)}=1, && \\
      & \widetilde{P}_{k,\ell}=0 \quad\text{if}\quad G_{k,\ell}=0, && \\
      & \widetilde{P}\geq 0,\quad b^{(0)}\geq 0 . &&
\end{alignat}
\end{subequations}

We solve this non-convex problem with a block coordinate descent method, alternating minimization of $\widetilde{P}$ and $b^{(0)}$. We initialize $\widetilde{P}$ and $b^{(0)}$ to be uniform. At each iteration we (i) calculate and fix $M$ and $b^{(0)}$, which makes Problem~\eqref{eq:optimization-problem} convex in $\widetilde{P}$ and solvable with CVXPY; (ii) recalculate $M$ with the result for $\widetilde{P}$; and (iii) take a projected gradient step on $b^{(0)}$. We find that running this for 100 iterations is sufficient to solve the objective function within an error of $10^{-3}$ for the tool distributions. This is a fixed amortized cost prior to generation.

\begin{figure}[ht]
\vskip 0.2in
\begin{center}
\centerline{\includegraphics[width=\columnwidth]{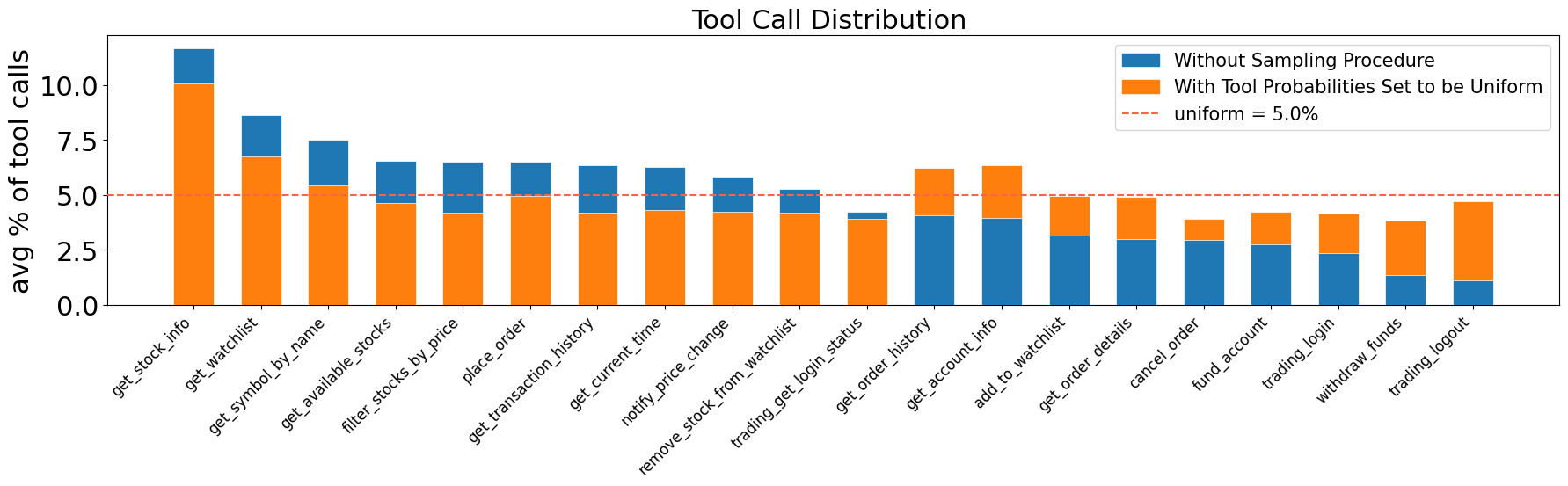}}
\caption{Distribution of tools in the generated data for the TradingBot API with and without the tool sampling procedure described.}
\label{fig:tc-distr}
\end{center}
\vskip -0.2in
\end{figure}

In Fig.~\ref{fig:tc-distr} we show the results from generation with the tool sampling procedure described above. We compare this with the case where all conditional tool probabilities are sampled uniformly across the states. The mean absolute error between the observed distribution of tools in the dataset and a uniform distribution is 42\% for the unoptimized case and 19\% for the optimized version. The residual error is partly attributable to the phantom tool calls, which are inserted into the tool sequence after sampling and skew the distribution away from uniform.

\section{Further Results}\label{app:further-results}
In this section, we present results of individual students (aggregated in the main body in Table \ref{tab:main-results}). The results of Qwen3.5-2B on TradingBot are lower than the results reported in Table \ref{tab:exps-comparison}. The experimental setup was different, but we will investigate the exact reason in future work. Nevertheless, the ranking of the methods holds.

\begin{table}[h!]
\centering
\small
\setlength{\tabcolsep}{5pt}
\begin{tabular}{l l cc cc cc cc cc}
\toprule
\multirow{2}{*}{Method} & \multirow{2}{*}{Model}
 & \multicolumn{2}{c}{Trading}
 & \multicolumn{2}{c}{Travel}
 & \multicolumn{2}{c}{VehicleControl}
 & \multicolumn{2}{c}{FileSystem$^1$}
 & \multicolumn{2}{c}{Tick.\&Trav.$^2$} \\
\cmidrule(lr){3-4} \cmidrule(lr){5-6} \cmidrule(lr){7-8}
\cmidrule(lr){9-10} \cmidrule(lr){11-12}
 & 
 & \makecell{Per\\Turn} & Full
 & \makecell{Per\\Turn} & Full
 & \makecell{Per\\Turn} & Full
 & \makecell{Per\\Turn} & Full
 & \makecell{Per\\Turn} & Full \\
\midrule
\multirow{2}{*}{Base Student}
 & Qwen  & 79.9 & 52.6 & 62.3 & 30.0 & 53.9 & 9.9  & 68.7 & 25.3 & 55.6 & 23.1 \\
 & Llama & 5.5  & 0.0  & 33.6 & 13.0 & 23.1 & 1.3  & 30.0 & 12.5 & 13.0 & 1.9  \\
\midrule
\multirow{2}{*}{Baseline}
 & Qwen  & 76.0 & 39.0 & 74.8 & 38.5 & 64.9 & 10.7 & 82.3 & 65.5 & 72.6 & 30.7 \\
 & Llama & 77.7 & 48.7 & 84.4 & 57.1 & 73.8 & 26.9 & 79.1 & 69.0 & 83.5 & 46.2 \\
\midrule
\multirow{2}{*}{Turns Mutator}
 & Qwen  & 73.8 & 38.3 & 76.5 & 42.3 & 64.1 & 13.1 & 89.2 & 72.4 & 69.6 & 30.8 \\
 & Llama & 78.2 & 51.4 & 76.5 & 42.3 & 64.4 & 25.0 & 88.6 & 75.9 & 82.7 & 50.0 \\
\midrule
\multirow{2}{*}{Markovian}
 & Qwen  & 73.9 & 34.2 & 77.7 & 46.2 & 69.6 & 14.3 & 93.0 & 79.3 & 68.2 & 23.1 \\
 & Llama & 79.6 & 48.8 & 88.5 & 75.0 & 70.9 & 25.0 & 96.8 & 89.7 & 83.5 & 50.0 \\
\midrule
\multirow{2}{*}{FSM}
 & Qwen  & 81.6 & 51.5 & 78.2 & 46.2 & 68.4 & 15.0 & 93.4 & 74.1 & 68.9 & 28.9 \\
 & Llama & 81.4 & 56.8 & 90.2 & 76.8 & 76.8 & 30.0 & 95.6 & 86.2 & 87.4 & 65.4 \\
\bottomrule
\end{tabular}
\caption{\centering Fine-tuning performance across generation conditions for both student
models, showing per-turn accuracy (\%) and full-conversation accuracy (\%).}
\flushleft
\hspace{10pt}\footnotesize $^1$ GorillaFileSystem API \hspace{5pt} $^2$ TicketAPI and TravelAPI
\label{tab:model-results}
\end{table}

\section{Training Details}\label{sec:training_details}

All models are fine-tuned with LoRA using AdamW as the optimizer for four epochs with a batch size of 1, initial learning rate of $5\cdot10^{-5}$ with linear learning rate decay, and LoRA rank 64. Each training run was performed on an Nvidia A100 GPU with synthetic data generation and finetuning taking approximately 72 hours per experiment.

\newpage
\section*{NeurIPS Paper Checklist}

\begin{enumerate}

\item {\bf Claims}
    \item[] Question: Do the main claims made in the abstract and introduction accurately reflect the paper's contributions and scope?
    \item[] Answer: \answerYes{} 
    \item[] Justification: The abstract and introduction claim a method for multi-turn data synthesis with an automatically constructed FSM, single-call trajectory generation, and improved token efficiency which matches the description in Sec.~\ref{sec:method} and the experimental results in Table~\ref{tab:exps-comparison} and Table~\ref{tab:main-results}.

\item {\bf Limitations}
    \item[] Question: Does the paper discuss the limitations of the work performed by the authors?
    \item[] Answer: \answerYes{} 
    \item[] Justification: Limitations are discussed in a dedicated section (Sec.~\ref{sec:limitations}) describing limited baseline comparisons owing to the costs, the test set being partially LLM-generated, re-implementing the baseline comparisons, and the FSM verification as future work.

\item {\bf Theory assumptions and proofs}
    \item[] Question: For each theoretical result, does the paper provide the full set of assumptions and a complete (and correct) proof?
    \item[] Answer: \answerNA{} 
    \item[] Justification: There are no theorems or proofs. Appendix~\ref{app:tool-sampling} derives a procedure for sampling tools with all assumptions stated inline. 

    \item {\bf Experimental result reproducibility}
    \item[] Question: Does the paper fully disclose all the information needed to reproduce the main experimental results of the paper to the extent that it affects the main claims and/or conclusions of the paper (regardless of whether the code and data are provided or not)?
    \item[] Answer: \answerYes{} 
    \item[] Justification: The method is fully specified in Sec.~\ref{sec:method} with prompts given in App.~\ref{app:prompts}, training details in App.~\ref{sec:training_details}, and the data splits in supplementary material.

\item {\bf Open access to data and code}
    \item[] Question: Does the paper provide open access to the data and code, with sufficient instructions to faithfully reproduce the main experimental results, as described in supplemental material?
    \item[] Answer: \answerNo{} 
    \item[] Justification: The code for this work is proprietary. However, the paper provides the complete generation prompts (App.~\ref{app:prompts}), the tool sampling procedure (App~\ref{app:tool-sampling}), training details in App.~\ref{sec:training_details}, and the data splits in supplementary material. The publicly available BFCL~v4 benchmark has been used extensively which also enables faithful re-implementation.

\item {\bf Experimental setting/details}
    \item[] Question: Does the paper specify all the training and test details (e.g., data splits, hyperparameters, how they were chosen, type of optimizer) necessary to understand the results?
    \item[] Answer: \answerYes{} 
    \item[] Justification: The experimental setup is given in Sec.~\ref{sec:exp-setup}, training hyperparameters in App.~\ref{sec:training_details} (including optimizer, learning rate, scheduler, epochs, batch size and LoRA rank), and the data splits in supplementary material.

\item {\bf Experiment statistical significance}
    \item[] Question: Does the paper report error bars suitably and correctly defined or other appropriate information about the statistical significance of the experiments?
    \item[] Answer: \answerYes{} 
    \item[] Justification: Table~\ref{tab:main-results} reports regression coefficients with standard errors and p-values from a multiple linear regression accounting for categorical control variables (generation condition, API domain, and student models) to capture variability across domains, folds and students.

\item {\bf Experiments compute resources}
    \item[] Question: For each experiment, does the paper provide sufficient information on the computer resources (type of compute workers, memory, time of execution) needed to reproduce the experiments?
    \item[] Answer: \answerYes{} 
    \item[] Justification: Compute details are provided in App.~\ref{sec:training_details}.

\item {\bf Code of ethics}
    \item[] Question: Does the research conducted in the paper conform, in every respect, with the NeurIPS Code of Ethics \url{https://neurips.cc/public/EthicsGuidelines}?
    \item[] Answer: \answerYes{} 
    \item[] Justification: The authors have reviewed the code of ethics and are confident that the work is in line with the code. 

\item {\bf Broader impacts}
    \item[] Question: Does the paper discuss both potential positive societal impacts and negative societal impacts of the work performed?
    \item[] Answer: \answerNA{} 
    \item[] Justification: This is foundational work on improving synthetic training data generation for tool calling models with no direct negative concerns beyond generic impacts of SLMs.
    
\item {\bf Safeguards}
    \item[] Question: Does the paper describe safeguards that have been put in place for responsible release of data or models that have a high risk for misuse (e.g., pre-trained language models, image generators, or scraped datasets)?
    \item[] Answer: \answerNA{} 
    \item[] Justification: the paper does not release scraped datasets or models.

\item {\bf Licenses for existing assets}
    \item[] Question: Are the creators or original owners of assets (e.g., code, data, models), used in the paper, properly credited and are the license and terms of use explicitly mentioned and properly respected?
    \item[] Answer: \answerYes{} 
    \item[] Justification: All models, baseline methods, and benchmarks are cited in the paper and used under their respective licenses.

\item {\bf New assets}
    \item[] Question: Are new assets introduced in the paper well documented and is the documentation provided alongside the assets?
    \item[] Answer: \answerYes{} 
    \item[] Justification: We release the domain-partitioned BFCL v4 train/test splits and the synthetic supplementary test trajectories as supplementary material with their use and generation method described in the paper. The data is released in line with the licensing terms from the models and datasets used to generate them.

\item {\bf Crowdsourcing and research with human subjects}
    \item[] Question: For crowdsourcing experiments and research with human subjects, does the paper include the full text of instructions given to participants and screenshots, if applicable, as well as details about compensation (if any)? 
    \item[] Answer: \answerNA{} 
    \item[] Justification: The paper does not involve crowdsourcing or human subject research.

\item {\bf Institutional review board (IRB) approvals or equivalent for research with human subjects}
    \item[] Question: Does the paper describe potential risks incurred by study participants, whether such risks were disclosed to the subjects, and whether Institutional Review Board (IRB) approvals (or an equivalent approval/review based on the requirements of your country or institution) were obtained?
    \item[] Answer: \answerNA{} 
    \item[] Justification: The paper does not involve human subject research.

\item {\bf Declaration of LLM usage}
    \item[] Question: Does the paper describe the usage of LLMs if it is an important, original, or non-standard component of the core methods in this research? Note that if the LLM is used only for writing, editing, or formatting purposes and does \emph{not} impact the core methodology, scientific rigor, or originality of the research, declaration is not required.
    \item[] Answer: \answerYes{} 
    \item[] Justification: LLMs are core components of the method and fully described in Sec.~\ref{sec:method}, Sec.~\ref{sec:exp-setup}, and App.~\ref{app:prompts}.

\end{enumerate}

\end{document}